# A Divergence Critic for Inductive Proof

**Toby Walsh** TOBY@ITC.IT
*IRST, Location Pantè di Povo*
*I38100 Trento, ITALY*

## Abstract

Inductive theorem provers often diverge. This paper describes a simple critic, a computer program which monitors the construction of inductive proofs attempting to identify diverging proof attempts. Divergence is recognized by means of a "difference matching" procedure. The critic then proposes lemmas and generalizations which "ripple" these differences away so that the proof can go through without divergence. The critic enables the theorem prover SPIKE to prove many theorems completely automatically from the definitions alone.

## 1. Introduction

Two key problems in inductive theorem proving are proposing lemmas and generalizations. A prover's divergence often suggests to the user an appropriate lemma or generalization that will enable the proof to go through without divergence. As a simple example, consider the theorem,

$$\forall n \,.\, dbl(n) \;=\; n + n.$$

This is part of a simple program verification problem (Dershowitz & Pinchover, 1990). Addition and doubling are defined recursively by means of the rewrite rules,

$$\begin{aligned} X + 0 &= X \\ X + s(Y) &= s(X + Y) \\ dbl(0) &= 0 \\ dbl(s(X)) &= s(s(dbl(X))) \end{aligned}$$

where $s(X)$ represents the successor of $X$ (that is, $X + 1$). I have adopted the PROLOG convention of writing meta-variables like $X$ and $Y$ in upper case.

The theorem prover SPIKE (Bouhoula, Kounalis, & Rusinowitch, 1992) fails to prove this theorem. The proof attempt begins with a simple one step induction on $x$. The base case is trivial. In the step case, the induction hypothesis is,

$$dbl(x) = x + x$$

And the induction conclusion is,

$$dbl(s(x)) = s(x) + s(x).$$

To ease presentation, variables in this paper are, as here, sometimes renamed from those introduced by SPIKE. This has no effect on the results as the prover and divergence critic both alpha convert variable names where necessary. Rewriting the induction conclusion with the recursive definitions of *dbl* and + gives,

$$s(s(dbl(x))) = s(s(x) + x).$$

The outermost successor functions on either side of the equality are now cancelled,

$$s(dbl(x)) = s(x) + x.$$

The prover then fertilizes with the induction hypothesis on the left hand side,

$$s(x + x) = s(x) + x.$$

This equation cannot be simplified further so another induction is performed. Unfortunately, this generates the diverging sequence of subgoals,

$$\begin{array}{rcl} s(x+x) & = & s(x)+x \\ s(s(x+x)) & = & s(s(x))+x \\ s(s(s(x+x))) & = & s(s(s(x)))+x \\ s(s(s(s(x+x)))) & = & s(s(s(s(x))))+x \\ s(s(s(s(s(x+x))))) & = & s(s(s(s(s(x)))))+x \\ & \vdots & \end{array}$$

The problem is that the prover repeatedly tries an induction on $x$ but is unable to simplify the successor functions that this introduces on the first argument position of +. The proof will go through without divergence if we have the rewrite rule,

$$s(X) + Y = s(X + Y).$$

This rule "ripples" accumulating successor functions off the first argument position of +. This rewrite rule is derived from the lemma,

$$\forall x, y \; . \; s(x) + y = s(x + y).$$

This is the commuted version of the recursive definition of addition and is, coincidently, a generalization of the first subgoal. This lemma can be proved without divergence as the induction variable, $y$ occurs just in the second argument position of +.

In this paper I describe a simple "divergence critic", a computer program which attempts to automate this process. The divergence critic identifies when a proof attempt is diverging by means of a "difference matching" procedure. The critic then proposes lemmas and generalizations which hopefully allow the proof to go through without divergence. Although the critic is designed to work with the prover Spike, it should also work with other induction provers (Walsh, 1994). Spike is a rewrite based theorem prover for first order conditional theories. It contains powerful rules for case analysis, simplification and implicit induction using the notion of a test set. Unfortunately, as is the case with other inductive theorem provers, its attempts to prove many theorems diverge without an appropriate generalization or the addition of a suitable lemma.

In Section 2, I describe difference matching and rippling, the two key ideas at the heart of the divergence critic. I then outline how difference matching identifies the accumulating term structure which is causing divergence (Section 3). In Section 4 and 6, I show how lemmas are speculated which "ripple" this term structure out of the way. In Section 5, I describe the heuristics used in generalizing these lemmas. Finally, implementation and results are described in Sections 7 and 8.

## 2. Difference matching and rippling

Rippling is a powerful heuristic developed at Edinburgh for proving theorems involving explicit induction (Bundy, Stevens, van Harmelen, Ireland, & Smaill, 1993) and is implemented in the Clam theorem prover (Bundy, van Harmelen, Horn, & Smaill, 1990). In the step case of an inductive proof, the induction conclusion typically differs from the induction hypothesis by the addition of some constructors or destructors. Rippling uses annotations to mark these differences and applies annotated rewrite rules to remove them.

As a simple example, consider again the theorem discussed in the introduction. In the step case, the induction hypothesis is,

$$dbl(x) = x + x$$

And the induction conclusion is,

$$dbl(s(x)) = s(x) + s(x).$$

If we "difference match" the induction conclusion against the induction hypothesis (Basin & Walsh, 1992), we obtain the following annotated induction conclusion,

$$dbl(\boxed{s(\underline{x})}) = \boxed{s(\underline{x})} + \boxed{s(\underline{x})}.$$

An annotation consists of a *wave-front*, a box with a *wave-hole*, an underlined term. Wave-fronts are always one functor thick (Basin & Walsh, 1994). That is, every wave-front has one immediate subterm that is annotated with a wave-hole. To make presentation simpler, we display adjacent wave-fronts merged. Thus, $\boxed{s(s(\underline{x}))}$ is just syntactic sugar for the annotated term, $\boxed{s(\underline{\boxed{s(\underline{x})}})}$. Wave-fronts can also include up and down arrows to indicate whether they are moving towards the top of the term tree or down towards the leaves. This extension can, however, be safely ignored here.

The *skeleton* of an annotated term is formed by deleting everything that appears in the wave-front but not in the wave-hole. The *erasure* of an annotated terms is formed by deleting the annotations but not the terms they contain. In this case, the skeleton of the annotated induction conclusion is identical to the induction hypothesis, and the erasure of the annotated induction conclusion is the unannotated induction conclusion. Difference matching guarantees this; that is, difference matching the induction conclusion with the induction hypothesis annotates the induction conclusion so that its skeleton matches the induction hypothesis.

Formally, $r$ is a *difference match* of $s$ with $t$ with substitution $\sigma$ iff $\sigma(skeleton(r)) = t$ and $erase(r) = s$ where $skeleton(r)$ and $erase(r)$ build the skeleton and erasure of the annotated term $r$. Difference matching is not unitary. That is, two terms can have more than one difference match. For example, both $\boxed{s(\underline{s(x)})}$ and $s(\boxed{s(\underline{x})})$ are difference matches of $s(s(x))$ with $s(x)$. The number of difference matches can be reduced if we compute just the *maximal* difference match in which wave-fronts are as high as possible in the term tree. A formal definition of such a well founded ordering on annotated terms has been given by Basin and Walsh (1994).

The aim of rippling is to rewrite the annotated induction conclusion so that the skeleton, the induction hypothesis, is preserved and the differences, the wave-fronts are moved to harmless places (for example, to the top of the term). If this rewriting succeeds, we will then be able to appeal to the induction hypothesis. To rewrite the annotated induction conclusion, we use the following annotated rewrite rules, or *wave rules*:

$$dbl(\boxed{s(\underline{X})}) = \boxed{s(s(\underline{dbl(X)}))} \quad (1)$$

$$X + \boxed{s(\underline{Y})} = \boxed{s(\underline{X+Y})} \quad (2)$$

$$\boxed{s(\underline{X})} + Y = \boxed{s(\underline{X+Y})} \quad (3)$$

The first two of these annotated rewrite rules are derived from the recursive definitions of *dbl* and + whilst the second is derived from the lemma proposed at the end of the introduction. Each of these annotated rewrite rules preserves the skeleton of the term being rewritten, and moves the wave-fronts higher up the term tree. Wave rules guarantee this: a wave rule is an annotated rewrite rule with an identical skeleton on left and right hand sides that moves wave-fronts in a well founded direction like, for instance, to the top of the term tree (Basin & Walsh, 1994).

Rippling on the left hand side of the annotated induction conclusion using (1) yields,

$$\boxed{s(s(\underline{dbl(x)}))} = \boxed{s(\underline{x})} + \boxed{s(\underline{x})}.$$

Then rippling on the right hand side with (2) gives,

$$\boxed{s(s(\underline{dbl(x)}))} = \boxed{s(\underline{\boxed{s(\underline{x})} + x})}.$$

Finally rippling with (3) on the right hand side yields,

$$\boxed{s(s(\underline{dbl(x)}))} = \boxed{s(s(\underline{x+x}))}.$$

As the wave-fronts are at the top of each term, we have successfully rippled both sides of the equality. We can now appeal to the induction hypothesis on the left hand side giving,

$$\boxed{s(s(\underline{x+x}))} = \boxed{s(s(\underline{x+x}))}.$$

This is a simple identity and the proof is complete. Note that to complete the proof, we needed to rewrite with a lemma, (3). The aim of the divergence critic described in this paper is to propose such lemmas.

Rippling has several desirable properties. It is highly goal directed, manipulating just the differences between the induction hypothesis and the induction conclusion. As the annotations restrict the application of the rewrite rules, rippling also involves little or no search. Difference matching and rippling have proved useful in domains outside of explicit induction. For example, they have been used to sum series (Walsh, Nunes, & Bundy, 1992) and to prove limit theorems (Yoshida, Bundy, Green, Walsh, & Basin, 1994). In the rest of the paper, I show that difference matching and rippling are also useful in identifying and correcting divergence in a prover that neither uses explicit rules of induction nor uses annotations to control rewriting.

## 3. Divergence Analysis

The initial problem is recognizing when the proof is diverging. Various properties of rewrite rules have been identified which cause divergence like, for example, forwards and backwards crossed systems (Hermann, 1989). However, these properties fail to capture all diverging rewrite systems since the problem is, in general, undecidable. The divergence critic instead studies the proof attempt looking for patterns of divergence; no attempt is made to analyse the rewrite rules themselves for structures which give rise to divergence. The advantage of this approach is that the critic need not know the details of the rewrite rules applied, nor the type of induction being performed, nor the control structure used by the prover. The critic can thus recognise divergence patterns arising from complex mutual or multiple inductions with little more difficulty than divergence patterns arising from simple straightforward inductions. The disadvantage of this approach is that the critic can identify a "divergence" pattern when none exists. Fortunately, such cases appear to be rare, and even when they occur, the critic usually suggests a lemma or generalization which gives a shorter and more elegant proof (see Section 8 for an example).

To illustrate the ideas behind the critic's divergence analysis, consider again the theorem from the introduction,

$$\forall n \ . \ dbl(n) \quad = \quad n + n.$$

The divergence critic first partitions the sequence of equations which the prover attempts to prove by induction. This is necessary since several diverging sequences may be interleaved in the prover's output. Several heuristics can be used to reduce the number of partitions considered. The most useful heuristic is parentage in which the sequence is partitioned so that each equation is derived from the previous one. That is, the equations lie on a single branch of the proof tree. In particular, the base case and step case of an induction are partitioned into different sequences. Other heuristics which can be used include: the function and constant symbols which occur in one equation occur in the next equation in the partition, and the weights of the equations in a partition form a simple arithmetic progression. In this case, there is just a single open branch in the proof tree,

$$\begin{aligned} s(x+x) &= s(x)+x \\ s(s(x+x)) &= s(s(x))+x \\ s(s(s(x+x))) &= s(s(s(x)))+x \\ &\vdots \end{aligned}$$

The critic then attempts to find the accumulating and nested term structure in each sequence which is causing divergence. In this case, successor functions are accumulating on the first argument of $+$. To identify this accumulating term structure, the critic uses difference matching. Difference matching successive equations gives the annotated sequence,

$$\begin{array}{rcl} s(x+x) & = & s(x)+x \\ \boxed{s(\underline{s(x+x)})} & = & \boxed{s(\underline{s(x)})}+x \\ \boxed{s(\underline{s(s(x+x))})} & = & \boxed{s(\underline{s(s(x))})}+x \\ & \vdots & \end{array}$$

This is the unique maximal difference match.

The critic then tries to speculate a lemma which can be used as a rewrite rule to move the accumulating and nested term structure out of the way. In this case, the critic speculates a rule for moving a successor function off the first argument of $+$. That is, the rule,

$$\boxed{s(\underline{X})}+Y = \boxed{s(\underline{X+Y})}.$$

With this rule, SPIKE is able to prove the *dbl* theorem without divergence. In addition, this rule is sufficiently simple that it can be proved without assistance. The heuristics used by the critic to perform this lemma speculation are described in more detail in the next two sections.

The divergence analysis performed by the critic is summarised in Figure 1. In analysing

1. There is a sequence of equations $s_i = t_i$ which the prover attempts to prove by induction ($i = 0,\ 1\ \ldots$);
2. There exists (non trivial) $G, H$ such that for each $j$, the maximal difference match has $s_j = G(U_j)$, and $s_{j+1} = G(\boxed{H(\underline{U_j})})$.

Figure 1: Preconditions to the divergence critic

the divergence, we consider *all* the equations which the prover attempts to prove by induction. This includes those equation where the induction proof succeeds as these can often suggest useful patterns. By "non-trivial" I wish to exclude $\lambda z.\, z$, the identity substitution. $H$ is thus the accumulating and nested term structure that appears to be causing divergence. For the *dbl* example, $H$ is $\lambda z.\, s(z)$, $G$ is $\lambda z.\, z + x$, and $U_0$ is $s(x)$. Although $G$ and $H$ are second order variables, the second order nature of the divergence analysis is limited. Indeed, the implementation of the critic merely requires first order difference matching which is polynomial. For simplicity, the preconditions ignore the orientation of equations. In addition, the preconditions can be easily generalised to include multiple and nested annotations. This allows the critic to recognise multiple sources of divergence in the same equation. Techniques which identify accumulating term structure by most specific generalization (Dershowitz & Pinchover, 1990) cannot cope with divergence patterns that give rise to nested annotations (see Section 9 for more details).

The specification of the preconditions has left the length of sequence undefined. If the sequence is of length 2, then the critic is preemptive. That is, it will propose a lemma just before another induction is attempted and divergence begins. Such a short sequence risks identifying divergence when none exists. On the other hand using a long sequence is expensive to test and allows the prover to waste time on diverging proof attempts. Empirically, a good compromise appears to be to look for sequences of length 3. This is both cheap to test and reliable. To identify accumulating term structure, it appears to be sufficient to use ground difference matching with alpha conversion of variable names. There exists a fast polynomial algorithm to perform such difference matching based upon the ground difference matching algorithm using dynamic programming (Basin & Walsh, 1993). Since the skeleton must be well typed (along with the erasure), the algorithm is extended to use sort information to prune potential difference matches.

## 4. Lemma Speculation

One way of removing the accumulating and nested term structure is to propose a wave rule which moves this difference to the top of the term leaving the skeleton unchanged. We hope either that it will then cancel against wave-fronts on the other side of the equality or that it will disappear in the process of being moved. For the *dbl* theorem, after generalization (which is discussed in the next section) the divergence pattern suggests a rule of the form,

$$\boxed{s(\underline{X})} + Y = \boxed{F(\underline{X+Y})}$$

where $F$ is a second order variable which we need to instantiate. Instantiating $F$ is ultimately a difficult synthesis problem so we can only hope to have heuristics that will work some of the time. Two of the heuristics used by the divergence critic to instantiate $F$ are *cancellation* and *petering out*.

The cancellation heuristic uses difference matching to identify term structure accumulating on the opposite side of the sequence which would allow cancellation to occur. Failing that, the cancellation heuristic looks for suitable term structure to cancel against in a new sequence (the original sequence is usually a divergence pattern of a step case, whilst the new sequence is usually a divergence pattern of a base case). In the *dbl* example, successor functions accumulate at the top of the left hand side of the diverging equations,

$$\begin{array}{rcl} s(x+x) & = & s(x)+x \\ \boxed{s(\underline{s(x+x)})} & = & \boxed{s(\underline{s(x)})}+x \\ \boxed{s(\underline{s(s(x+x))})} & = & \boxed{s(\underline{s(s(x))})}+x \\ & \vdots & \end{array}$$

This divergence pattern suggests that $F$ should be instantiated to $\lambda z.\, s(z)$ to enable immediate cancellation. Thus, as required, the cancellation heuristic suggests the rule,

$$\boxed{s(\underline{X})} + Y = \boxed{s(\underline{X + Y})}.$$

The other heuristic used to instantiate the right hand side of speculated lemmas is petering out. In moving the differences up to the top of the term, they may disappear altogether. Consider, for example, the theorem,

$$\forall l \;.\; sorted(isort(l)) = true$$

where *isort* is insertion sort and *sorted* is true iff a list is sorted into order. These are defined by the conditional rewrite rules,

$$\begin{aligned} sorted(nil) &= true \\ sorted(cons(X, nil)) &= true \\ X < Y \rightarrow sorted(cons(X, cons(Y, Z))) &= sorted(cons(Y, Z)) \\ isort(nil) &= nil \\ isort(cons(X, Y)) &= insert(X, isort(Y)) \end{aligned}$$

where $insert(X, Z)$, which inserts the element $X$ into the list $Z$ in order, and $X < Y$ are defined by the rewrite rules,

$$\begin{aligned} 0 < X &= true \\ s(X) < 0 &= false \\ s(X) < s(Y) &= X < Y \\ insert(X, nil) &= cons(X, nil) \\ X < Y \rightarrow insert(X, cons(Y, Z)) &= cons(X, cons(Y, Z)) \\ \neg(X < Y) \rightarrow insert(X, cons(Y, Z)) &= cons(Y, insert(X, Z)) \end{aligned}$$

Divergence analysis of SPIKE's attempt to prove this theorem suggests the need for a rule of the form,

$$sorted(\boxed{insert(Y, \underline{X})}) = \boxed{F(\underline{sorted(X)})}.$$

The petering out heuristic instantiates $F$ to the identity function $\lambda z.\, z$. This gives the rule,

$$sorted(\boxed{insert(Y, \underline{X})}) = sorted(X).$$

This rule allows the proof to go through without divergence.

As a more complex example, consider the theorem,

$$\forall i, j, l \;.\; nth(i, nth(j, l)) = nth(j, nth(i, l))$$

where $nth$ is defined by the rewrite rules,

$$\begin{aligned} nth(0, L) &= L \\ nth(N, nil) &= nil \\ nth(s(N), cons(H, T)) &= nth(N, T). \end{aligned}$$

SPIKE's diverging attempt to prove this theorem generates the equations,

$$\begin{array}{rcl} nth(s(i), nth(j, x)) & = & nth(s(j), nth(i, x)) \\ nth(s(s(i)), nth(j, cons(y, x))) & = & nth(s(j), nth(i, x)) \\ nth(s(s(s(i))), nth(j, cons(z, cons(y, x)))) & = & nth(s(j), nth(i, x)) \\ & \vdots & \end{array}$$

Divergence analysis identifies term structure accumulating in two different places,

$$\begin{array}{rcl} nth(s(i), nth(j, x)) & = & nth(s(j), nth(i, x)) \\ nth(\boxed{s(\underline{s(i)})}, nth(j, \boxed{cons(y, \underline{x})})) & = & nth(s(j), nth(i, x)) \\ nth(\boxed{s(\underline{s(s(i))})}, nth(j, \boxed{cons(z, \underline{cons(y, x)})})) & = & nth(s(j), nth(i, x)) \\ & \vdots & \end{array}$$

This is the unique maximal difference match. This divergence pattern suggests the need for a rewrite rule of the form,

$$nth(\boxed{s(\underline{I})}, nth(J, \boxed{cons(Y, \underline{X})})) = \boxed{F(\underline{nth(I, nth(J, X))})}.$$

The petering out heuristic instantiates $F$ to the identity function $\lambda z.\, z$ giving the rule,

$$nth(\boxed{s(\underline{I})}, nth(J, \boxed{cons(Y, \underline{X})})) = nth(I, nth(J, X)).$$

This rule allows the proof to go through without divergence.

Since the erasure of the wave rule must be properly typed, sort information can be used to prune inappropriate instantiations for $F$. All speculated lemmas are therefore filtered through a type checker. Speculated lemmas are also filtered through a conjecture disprover. When a confluent set of rewrite rules exists for ground terms, exhaustive normalization of some representative set of ground instances of the equations is used to filter out non-theorems. Alternatively, the prover itself could be used to filter out non-theorems. Unlike many other induction theorem provers, SPIKE can refute conjectures since its inference rules are refutationally complete for conditional theories in which the axioms are ground convergent and defined functions are completely defined over free constructors (Bouhoula & Rusinowitch, 1995a). Other techniques for disproving conjectures are described by Protzen (1992).

The critic's lemma speculation is summarized in Figure 2 (using the same variable names as the preconditions). This specification again uses second order variables in a limited manner. First order difference matching is merely required to construct lemmas. As with the preconditions, the specification of the postconditions can be easily extended to deal with multiple and nested wave-fronts (as in the $nth(i, nth(j, l)) = nth(j, nth(i, l))$ example). Since the rules proposed by the critic move the wave-fronts to top of the term, they usually only introduce fresh divergence in the rare cases that cancellation or fertilization fails. This is unlikely since the cancellation and petering out heuristics attempt to ensure *precisely* that cancellation or fertilization can take place.

1. The critic proposes a rule of the form,

$$G(\boxed{H(\underline{U_0})}) \quad = \quad \boxed{F(\underline{G(U_0)})}$$

2. $F$ is instantiated by the cancellation or petering out heuristics;

3. Lemmas are filtered through a type checker and a conjecture disprover;

4. If several lemmas are suggested, the critic deletes any that are subsumed.

Figure 2: Postconditions to the divergence critic

## 5. Generalization

A major cause of divergence is the need to generalize. Most of the lemmas proposed by the critic fix divergence, but attempting to prove the lemmas themselves can cause fresh divergence. In addition, several speculated lemmas can sometimes be replaced by a single generalization. Generalized lemmas also can lead to shorter, more elegant and natural proofs. The critic therefore attempts to generalize the lemma speculated, using the conjecture disprover to guard against over-generalization.

The main heuristic used for generalization is an extension of the primary term heuristic (Aubin, 1976). The *primary terms* are those terms encountered as a term is explored from the root to the leaves ignoring non-recursive argument positions to functions. The same notion of recursive argument position is used by the critic as defined by Bouhoula and Rusinowitch (1995a) and as used by SPIKE for performing inductions.

Consider, for example, the theorem,

$$\forall a, b \;.\; len(a) + len(b) = len(app(a, b))$$

where $+$ is again defined recursively on its second argument, and *len* and *app* are defined by means of the rewrite rules,

$$\begin{aligned} len(nil) &= 0 \\ len(cons(H, T)) &= s(len(T)) \\ app(nil, T) &= T \\ app(cons(H, T), R) &= cons(H, app(T, R)). \end{aligned}$$

This problem is taken from the CLAM library corpus (Bundy et al., 1990). SPIKE's attempt to prove this theorem diverges. One of the sequences of equations generated is,

$$0 + len(b) \quad = \quad len(b)$$

$$\begin{aligned} s(0) + len(b) &= s(len(b)) \\ s(s(0)) + len(b) &= s(s(len(b))) \\ &\vdots \end{aligned}$$

Difference matching identifies the term structure causing divergence,

$$\begin{aligned} 0 + len(b) &= len(b) \\ \boxed{s(\underline{0})} + len(b) &= \boxed{s(\underline{len(b)})} \\ \boxed{s(\underline{s(0)})} + len(b) &= \boxed{s(\underline{s(len(b))})} \\ &\vdots \end{aligned}$$

This is the unique maximal difference match. These annotations suggest the need for the wave rule,

$$\boxed{s(\underline{0})} + len(B) = \boxed{s(\underline{0 + len(B)})}.$$

A set of candidate terms for generalization is constructed by computing the intersection of the primary terms of the two sides of this rule. In this case, the primary terms of the left hand side are the set $\{s(0) + len(B), len(B), B\}$, and the primary terms of the right hand side are the set $\{s(0 + len(B)), 0 + len(B), len(B), B\}$. The intersection of the primary terms is thus the set $\{len(B), B\}$. The critic picks members of the intersection to generalize to new variables. Picking $B$ justs gives an equivalent lemma up to renaming of variables. Picking $len(B)$ gives the generalization,

$$\boxed{s(\underline{0})} + Y = \boxed{s(\underline{0 + Y})}.$$

The reason for considering just primary terms is that the recursive definitions typically provide wave rules for removing term structure which accumulates at these positions. In addition to primary terms, the divergence critic therefore also considers the positions of the wave-holes (but not wave-fronts) in the skeleton of the lemma being speculated. The motivation for this extension is that the speculated lemma will allow accumulating term structure to be moved from the wave-hole positions; such positions are therefore also candidates for generalization. Positions of the wave-fronts are not included since we want to speculate a lemma that will move the term structure at such positions.

For instance, because of the wave-hole on the first argument of $+$ in the last example, 0 is also included in the intersection set of candidate terms for generalization. Picking 0 to generalize gives,

$$\boxed{s(\underline{X})} + Y = \boxed{s(\underline{X + Y})}.$$

The speculated lemma is now as general as is possible. This rule allows the proof to go through without divergence.

The critic also has a heuristic for merging speculated lemmas. For instance, with the theorem $sorted(isort(x))$, the critic speculates several rules including,

$$\begin{aligned} sorted(\boxed{insert(0, \underline{X})}) &= sorted(X) \\ sorted(\boxed{insert(s(Y), \underline{X})}) &= sorted(X) \end{aligned}$$

As $\{0, s(Y)\}$ is a cover set for the natural numbers, these two rules can be merged to give,

$$sorted(\boxed{insert(Y, \underline{X})}) = sorted(X).$$

## 6. Transverse Wave Rules

The lemmas speculated so far have moved accumulating term structure directly to the top of the term where it is removed by cancellation or petering out. An alternative way of removing accumulating term structure is to move it onto another argument position where: either it can be removed by matching with a "sink", a universally quantified variable in the induction hypothesis; or it can be moved upwards by rewriting with the recursive definitions. Annotated rewrite rules which preserve the skeleton and move wave-fronts across to other argument positions are called *transverse* wave rules (Bundy et al., 1993). Theorems involving functions with accumulators provide a rich source of examples where such rewrite rules prevent divergence.

Consider, for example, a theorem about the correctness of tail recursive list reversal,

$$\forall a, b \,.\, qrev(a, b) = app(rev(a), b)$$

where both $a$ and $b$ are universally quantified, $rev$ is naive list reversal using append, and $qrev$ is tail recursive list reversal building the reversed list on the second argument position. These functions are defined by the rewrite rules,

$$\begin{aligned} rev(nil) &= nil \\ rev(cons(H, T)) &= app(rev(T), cons(H, nil)) \\ qrev(nil, R) &= R \\ qrev(cons(H, T), R) &= qrev(T, cons(H, R)). \end{aligned}$$

Spike's attempt to prove this theorem diverges generating the following sequence of equations which the prover attempts to show by induction,

$$\begin{aligned} qrev(a, b) &= app(rev(a), b) \\ qrev(a, cons(c, b)) &= app(app(rev(a), cons(c, nil)), b) \\ qrev(a, cons(c, cons(d, b))) &= app(app(app(rev(a), cons(c, nil)), cons(d, nil)), b) \\ &\vdots \end{aligned}$$

Difference matching identifies the term structure accumulating within these equations that is causing divergence,

$$\begin{aligned}
qrev(a, b) &= app(rev(a), b) \\
qrev(a, \boxed{cons(c, \underline{b})}) &= app(\boxed{app(\underline{rev(a)}, cons(c, nil))}, b) \\
qrev(a, \boxed{cons(c, \underline{cons(d, b)})}) &= app(\boxed{app(\underline{app(rev(a), cons(c, nil))}, cons(d, nil))}, b) \\
&\vdots
\end{aligned}$$

This is the unique maximal difference match. Rather than move the accumulating term structure on the right hand side of the equations to the top of the term, it is much simpler to move the accumulating term structure from the first onto the second argument of the outermost append. The critic therefore proposes a transverse wave rule, which preserves the skeleton but moves the difference onto a different argument position. In this example, this is a rule of the form,

$$app(\boxed{app(\underline{rev(A)}, cons(C, nil))}, B) = app(rev(A), \boxed{F(\underline{B})}).$$

In moving the difference onto another argument position, the difference may change syntactically. The right hand side of the lemma is therefore only partially determined. To instantiate $F$, the critic uses two heuristics: *fertilization* and *simplification*.

The fertilization heuristic uses matching to find an instantiation for $F$ which enables immediate fertilization. In this case, matching against the universally quantified variable $b$ in the induction hypothesis suggests,

$$app(\boxed{app(\underline{rev(A)}, cons(C, nil))}, B) = app(rev(A), \boxed{cons(C, \underline{B})}).$$

Finally the critic generalizes the lemma using the same extended primary term heuristic as before (*i.e.*, augmenting recursive positions with wave-hole positions). This gives the rule,

$$app(\boxed{app(\underline{A}, cons(C, nil))}, B) = app(A, \boxed{cons(C, \underline{B})}).$$

This is exactly the rule needed by Spike to complete the proof. In addition, it is simple enough to be proved by itself without divergence; this is not true of the ungeneralized rule.

The other heuristic used to instantiate the right hand side of the speculated lemma is the simplification heuristic. The heuristic uses regular matching to find an instantiation for $F$ which will enable the wave-front to be simplified using one of the recursive definitions. Consider again the *dbl* theorem from the introduction. Divergence analysis identifies successor functions accumulating on the first argument position of $+$. This accumulating term structure can either be moved to the top of the term tree or alternatively onto the second argument position of $+$ using a transverse wave rule of the form,

$$\boxed{s(\underline{X})} + Y = X + \boxed{F(\underline{Y})}.$$

The right hand side of this transverse wave rule is instantiated by the simplification heuristic. The wave-front on the right hand side can be simplified by the rewrite rule recursively defining $+$ if $F$ is instantiated by $\lambda z \,.\, s(z)$. That is, if we have the rule,

$$\boxed{s(\underline{X})} + Y = X + \boxed{s(\underline{Y})}.$$

This rule allows the proof to go through without divergence.

Speculated transverse wave rules are generalized using the extended primary terms heuristic described in Section 5. The divergence critic also generalizes transverse wave rules by means of an *equality* heuristic. This heuristic attempts to cancel equal outermost functors where possible. For example, consider the theorem,

$$\forall x, y \,.\, (x + y) - x = y$$

where addition is defined recursively on its second argument position and subtraction is defined by the rewrite rules,

$$\begin{aligned} X - 0 &= X \\ 0 - X &= 0 \\ s(X) - s(Y) &= X - Y. \end{aligned}$$

SPIKE's attempt to prove this theorem diverges generating (amongst others) the goals,

$$\begin{aligned} (x + y) - x &= y \\ (s(x) + y) - x &= s(y) \\ (s(s(x)) + y) - x &= s(s(y)) \\ &\vdots \end{aligned}$$

Divergence analysis identifies accumulating term structure within these equations,

$$\begin{aligned} (x + y) - x &= y \\ (\boxed{s(\underline{x})} + y) - x &= \boxed{s(\underline{y})} \\ (\boxed{s(\underline{s(x)})} + y) - x &= \boxed{s(\underline{s(y)})} \\ &\vdots \end{aligned}$$

This is the unique maximal difference match. These annotations suggest the need for the transverse rule,

$$(\boxed{s(\underline{X})} + Y) - X = (X + \boxed{s(\underline{Y})}) - X.$$

The equality heuristic deletes the equal outermost function, $\lambda z \,.\, z - X$. This gives the more general lemma,

$$\boxed{s(\underline{X})} + Y = X + \boxed{s(\underline{Y})}.$$

All speculated lemmas are filtered through a type checker to ensure that their erasure is well typed. Speculated lemmas are also filtered through a conjecture disprover to guard against over-generalization.

The actions of the critic are summarized in Figure 3. The specification of preconditions and postconditions again uses second order variables but in a limited manner. The implementation merely requires second order matching and first order difference matching. The preconditions and postconditions can be easily generalised to include multiple and nested

Preconditions:

1. There is a sequence of equations $s_i = t_i$ which the prover attempts to prove by induction ($i = 0, 1 ...$);
2. There exists (non trivial) $G, H$ such that for each $j$, the maximal difference match has $s_j = G(U_j, Acc)$ and $s_{j+1} = G(\boxed{H(\underline{U_j})}, Acc)$.

Postconditions:

1. The critic proposes a rule of the form,

$$G(\boxed{H(\underline{U_0})}, Acc) \quad = \quad G(U_0, \boxed{F(\underline{Acc})})$$

2. $F$ is instantiated by the fertilization or simplification heuristics;
3. The lemma is generalized using the (augmented) primary terms and equality heuristics;
4. Generalized lemmas are filtered through a type checker and a conjecture disprover;
5. If several lemmas are suggested, the critic deletes any that are subsumed.

Figure 3: Speculation of transverse wave rules.

annotations. We could also speculate hybrid wave rules which ripple part of the wave-front across and part of it up the term tree. However, such rules appear to be rare. In addition, such hybrid wave rules can often be decomposed into a pair of wave rules, one of which moves some of the wave-fronts up the term tree, and another which moves the wave-fronts across.

## 7. Implementation

The divergence critic described in the previous sections has been implemented in Prolog. The system consists of 787 lines of code defining approximate 100 different Prolog predicates. More recently a cut down version has been incorporated directly within the Spike system which is written in Caml Light (Bouhoula & Rusinowitch, 1995b). The output of Spike is parsed to generate input to the critic. The input consists of: the equations which the prover attempts to prove by induction; sort information (for the type checker and difference matcher); the recursive argument positions (for constructing primary terms); and the rewrite rules defining the theory (used by the conjecture disprover).

```
% compiling file /home/dream5/tw/work/Spike/diverge/data.double.x+x
% data.double.x+x compiled in module user, 0.233 sec 1,612 bytes

| ?- speculate.

  Equations input:
  double(x1)=x1+x1
  s(x1+x1)=s(x1)+x1
  s(s(x1+x1))=s(s(x1))+x1
  s(s(s(x1+x1)))=s(s(s(x1)))+x1

  Lemmas speculated:
  s(x1)+x1=s(x1+x1)
  s(x1)+x99=s(x1+x99)
  s(x99)+x1=s(x99+x1)
  s(x99)+x100=s(x99+x100)
  s(x1)+x1=x1+s(x1)
  s(x1)+x99=x1+s(x99)
  s(x99)+x1=x99+s(x1)
  s(x99)+x100=x99+s(x100)

  Deleting lemmas subsumed:
  s(x1)+x99=s(x1+x99)
  s(x1)+x99=x1+s(x99)

  Merging remaining lemmas:
  s(x1)+x99=s(x1+x99)
  s(x1)+x99=x1+s(x99)

yes
| ?-
```

Figure 4: Example output of the divergence critic.

Figure 1 gives the divergence critic's output on the problem discussed in the introduction. Either of the proposed lemmas when used as a rewrite rule is adequate to fix divergence. In addition, the proposed lemmas are sufficiently simple to be proved automatically without introducing fresh divergence. The first lemma is a rewrite rule for moving accumulating successor functions from the first argument position of + to the top of the term tree. The second lemma is a transverse wave rule discussed in Section 6 for moving accumulating successor functions from the first argument position of + to the second argument position.

The critic is successful at identifying divergence and proposing appropriate lemmas and generalizations for a significant number of theorems. Divergence analysis is very quick on most examples. The divergence pattern is recognized usually in less than a second. Most of the time is spent looking for generalizations and refuting over-generalizations with the conjecture disprover. This usually takes between 1 and 100 seconds. Additional heuristics for preventing over-generalization and a more efficient implementation of the conjecture disprover would speed up the critic considerably.

## 8. Results

Table 1 lists 30 theorems that cause Spike to diverge and the lemmas speculated by the divergence critic after analysing the diverging proof attempts. These problems provide a representative sample of the type of theorems for which the cause of divergence can be identified and an appropriate lemma or generalization speculated. Many of these problems come from the Clam library corpus. Part of this table has appeared before (Walsh, 1994). Times are for the divergence critic to speculate the lemmas and are for the average of 10 runs on a Sun 4 running Quintus 3.1.1.

Spike's proof attempt diverges on each example when given the definitions alone. In each of the 30 cases, the critic is quickly able to suggest a lemma which overcomes divergence. When multiple lemmas are proposed (with the exception of 20) any one on its own is sufficient to fix divergence. In every case (except 13 and 24) the lemmas proposed are sufficiently simple to be proved automatically without introducing fresh divergence. In the majority of cases, the lemmas proposed are optimal; that is, they are the simplest possible lemmas which fix divergence. In the cases when the lemma is not optimal, they are usually only slightly more complex than the simplest lemma which fixes divergence. In many of the examples, other lemmas are conjectured by the divergence analysis but these are quickly rejected by the conjecture disprover. For example, in example 16, divergence analysis and the petering out heuristic suggest the rule,

$$\#\# \qquad len(\boxed{app(\underline{X}, cons(Y, nil))}) = len(X) \qquad \#\#$$

However, this is refuted by exhaustive normalization using any ground terms for $X$ and $Y$. In this case, the cancellation heuristic identifies the required lemma,

$$len(\boxed{app(\underline{X}, cons(Y, nil))}) = \boxed{s(\underline{len(X)})}.$$

Some of the examples deserve additional comment. In example 1, the divergence critic identifies that successor functions are accumulating on the first argument position of $+$. The critic speculates a lemma for moving these successor functions either to the top of the term (so that immediate cancellation can occur) or onto to the second argument position (so that simplification with the recursive definition of $+$ can occur). The first lemma speculated is in fact a generalization of the theorem being proved. Example 2 is a simple program verification problem taken from Dershowitz and Pinchover (1990). The forward direction of this theorem was discussed in the introduction. Similar divergence occurs as in example 1 and, after generalization, the same lemmas are speculated.

Example 3 caused divergence in the beta-version of Spike available in the summer of 1994. The proof rules in Spike have since been strengthened and this example no longer

| No | Theorem | Lemmas speculated | Time/s |
|---|---|---|---|
| 1 | s(x)+x=s(x+x) | s(X)+Y=s(X+Y)<br>s(X)+Y=X+s(Y) | 7.8 |
| 2 | dbl(x)=x+x ↔<br>dbl(0)=0, dbl(s(x))=s(s(dbl(x))) | s(X)+Y=s(X+Y)<br>s(X)+Y=X+s(Y) | 8.2 |
| 3 | len(x @ y)=len(y @ x) | len(X @ (Z :: Y))=s(len(X @ Y))<br>len(X @ (Z :: Y))=len((W :: X) @ Y) | 3.6 |
| 4 | len(x @ y)=len(x)+len(y) | s(X)+Y=s(X+Y)<br>s(X)+Y=X+s(Y) | 7.2 |
| 5 | len(x @ x)=dbl(len(x)) | len(X @ (Z :: Y))=s(len(X @ Y))<br>len(X @ (W :: Z :: Y))=s(len(X @ (W :: Y))) | 11.6 |
| 6 | even(x+x) | even(s(s(X))+Y)=even(X+Y) | 5.4 |
| 7 | odd(s(x)+x) | odd(s(s(X))+Y)=odd(X+Y) | 16.0 |
| 8 | even$_m$(x+x) | even$_m$(s(s(X))+Y)=even$_m$(X+Y)<br>odd$_m$(s(s(X))+Y)=odd$_m$(X+Y) | 28.4 |
| 9 | odd$_m$(s(x)+x) | even$_m$(s(s(X))+Y)=even$_m$(X+Y)<br>odd$_m$(s(s(X))+Y)=odd$_m$(X+Y) | 65.5 |
| 10 | even$_m$(x) → half(x)+half(x)=x | s(X)+Y=s(X+Y)<br>s(X)+Y=X+s(Y) | 6.0 |
| 11 | half(x+x)=x | s(s(X))+Y=X+s(s(Y))<br>half(s(s(X))+Y)=half(X+Y) | 11.1 |
| 12 | half(s(x)+x)=x | s(s(X))+Y=X+s(s(Y))<br>half(s(s(X))+Y)=half(X+Y) | 31.0 |
| 13 | rot(len(x),x)=x | rot(len(X),X @ [Y])=Y :: rot(len(X),X) | 2.4 |
| 14 | len(rot(len(x),x))=len(x) | len(rot(X,Z @ [Y]))=s(len(rot(X,Z))) | 4.8 |
| 15 | rot(len(x),x @ [y])=y :: rot(len(x),x) | (X @ [Y]) @ Z=X @ (Y :: Z)<br>rot(len(X),(X @ [Y]) @ Z)=Y :: rot(len(X),X @ Z) | 86.3 |
| 16 | len(rev(x))=len(x) | len(X @ [Y])=s(len(X)) | 2.0 |
| 17 | rev(rev(x))=x | rev(X @ [Y])=Y :: rev(X) | 1.2 |
| 18 | rev(rev(x) @ [y])=y :: x | rev(X @ [Y])=Y :: rev(X) | 16.0 |
| 19 | rev(rev(x) @ [y])=y :: rev(rev(x)) | rev(X @ [Y])=Y :: rev(X) | 18.6 |
| 20 | len(rev(x @ y))=len(x)+len(y) | len(X @ [Y])=s(len(X))<br>s(X)+Y=s(X+Y)<br>s(X)+Y=X+s(Y) | 10.0 |
| 21 | len(qrev(x,[]))=len(x) | len(qrev(X,Z :: Y))=s(len(qrev(X,Y))) | 2.2 |
| 22 | qrev(x,y)=rev(x) @ y | (X @ [Y]) @ Z=X @ (Y :: Z) | 3.4 |
| 23 | len(qrev(x,y))=len(x)+len(y) | s(X)+Y=s(X+Y)<br>s(X)+Y=X+s(Y) | 12.0 |
| 24 | qrev(qrev(x,[]),[])=x | qrev(qrev(X,[Y]),Z)=Y :: qrev(qrev(X,[]),Z) | 5.0 |
| 25 | rev(qrev(x,[]))=x | rev(qrev(X,[Y]))=Y :: rev(qrev(X,[])) | 5.8 |
| 26 | qrev(rev(x),[])=x | qrev(X @ [Y],Z)=Y :: qrev(X,Z) | 5.2 |
| 27 | nth(i,nth(j,x))=nth(j,nth(i,x)) | nth(s(I),nth(J,Y :: X))=nth(I,nth(J,X)) | 7.4 |
| 28 | nth(i,nth(j,nth(k,x)))=nth(k,nth(j,nth(i,x))) | nth(s(I),nth(J,Y :: X))=nth(I,nth(J,X)) | 7.6 |
| 29 | len(isort(x))=len(x) | len(insert(Y,X))=s(len(X)) | 2.0 |
| 30 | sorted(isort(x)) | sorted(insert(Y,X))=sorted(X)<br>sorted(insert(Y,insert(Z,X)))=sorted(X) | 114 |

Table 1: Some lemmas speculated by the divergence critic.

**Notes:** :: is written for infix cons, @ for infix append, [] for nil, and [$x$] for cons($x$,nil). In addition, even is defined by a s(s(x)) recursion, even$_m$ by a mutual recursion with odd$_m$, and rot$(n, l)$ rotates a list $l$ by $n$ elements.

causes divergence in the current release. The speculated lemmas do, however, simplify the proof. Example 4 was used in the text to illustrate the generalization heuristics. The second lemma in example 5 is perhaps a little surprising,

$$len(app(X,(cons(W,\boxed{cons(Z,\underline{Y})})))) = \boxed{s(\underline{len(app(X,cons(W,Y)))})}.$$

Although it is more complex than the first lemma, it is nearly as good at fixing divergence.

In example 6, the lemma proposed,

$$even(\boxed{s(s(\underline{X}))}+Y) = even(X+Y)$$

is not optimal. That is, it is not the simplest possible lemma that fixes divergence. To fix divergence, we merely need one of the rules, $s(X)+Y = s(X+Y)$ or $s(X)+Y = X+s(Y)$. Either of these will ripple the successor functions accumulating on the first argument position of $+$. The divergence critic attempts to construct a lemma to ripple two successor functions across from the first to the second argument positions of $+$. Unfortunately, the critic fails to find an appropriate instantiation for the right hand side of such a lemma. The critic instead proposes a rule to move the two successor functions up to the top of the term where the wave-front can peter out. Example 7 is very similar to example 6.

Examples 8 to 10 demonstrate that the critic can cope with divergence in theories involving mutual recursion. In example 8, Spike attempts to prove by induction the equations,

$$\begin{aligned}
even_m(x+x) &= true\\
odd_m(s(x)+x) &= true\\
even_m(s(s(x))+x) &= true\\
odd_m(s(s(s(x)))+x) &= true\\
even_m(s(s(s(s(x))))+x) &= true\\
&\vdots
\end{aligned}$$

The critic identifies two inter-linking divergence patterns,

$$\begin{array}{rl}
even_m(x+x) = true & odd_m(s(x)+x) = true\\
even_m(\boxed{s(s(\underline{x}))}+x) = true & odd_m(\boxed{s(s(\underline{s(x)}))}+x) = true\\
even_m(\boxed{s(s(\underline{s(s(x))}))}+x) = true & odd_m(\boxed{s(s(\underline{s(s(s(x)))}))}+x) = true\\
\vdots & \vdots
\end{array}$$

The critic therefore proposes rules which ripple this accumulating term structure up to the top of the term where it peters out,

$$\begin{aligned}
even_m(\boxed{s(s(\underline{X}))}+Y) &= even_m(X+Y)\\
odd_m(\boxed{s(s(\underline{X}))}+Y) &= odd_m(X+Y)
\end{aligned}$$

Just as in examples 6 and 7, these are not the optimal rules for fixing divergence. Nevertheless, either of the proposed rules fix divergence and both can be proved without difficulty by SPIKE. Example 9 is very similar to example 8.

Examples 10 to 12 require little comment. In example 13, the proposed lemma is too difficult to be proved automatically. However, the divergence critic is able to identify the cause of this difficulty and propose a lemma which allows the proof to go through (example 15). In example 14, the speculated lemma is not optimal. The simpler lemma speculated in example 13 would be adequate to prove this theorem without divergence. The speculated lemma is not optimal because the divergence critic attempts to ripple the accumulating term structure over two functors, *len* and *rot* to the top of the term tree. However, it is sufficient on this problem to ripple it up over just one functor, *rot*.

Examples 16 to 19 are straightforward and do not require discussion. In example 20, the critic identifies two separate divergence patterns. To overcome divergence, the first lemma plus one or other of the second and third are therefore needed. The first divergence pattern occurs in the sequence of subgoals,

$$\begin{aligned} len(rev(x)) &= 0 + len(x) \\ len(\boxed{app(\underline{rev(x)}, cons(y, nil))}) &= \boxed{s(\underline{0 + len(x)})} \\ len(\boxed{app(\underline{app(rev(x), cons(y, nil))}, cons(z, nil))}) &= \boxed{s(\underline{s(0 + len(x))})} \\ &\vdots \end{aligned}$$

Term structure is accumulating on the second argument of append. Such term structure is removed by the first rule,

$$len(\boxed{app(\underline{X}, cons(Y, nil))}) = \boxed{s(\underline{len(X)})}$$

The second divergence pattern occurs in the sequence of subgoals,

$$\begin{aligned} s(x) + len(y) &= s(x + len(y)) \\ \boxed{s(\underline{s(x)})} + len(y) &= \boxed{s(\underline{s(x + len(y))})} \\ \boxed{s(\underline{s(s(x))})} + len(y) &= \boxed{s(\underline{s(s(x + len(y)))})} \\ &\vdots \end{aligned}$$

Term structure is accumulating on the first argument of +. This is removed by one or other of the second and third rules,

$$\begin{aligned} \boxed{s(\underline{X})} + Y &= \boxed{s(\underline{X + Y})} \\ \boxed{s(\underline{X})} + Y &= X + \boxed{s(\underline{Y})} \end{aligned}$$

Examples 21 and 23 are reasonably straightforward. The lemma speculated in example 22 is a special case of the associativity of append. More powerful generalization heuristics could have speculated the associativity of append. However, such heuristics would also speculate more non-theorems. Further research into the optimal strength of generalization heuristics would be valuable.

Example 24 is the only disappointment; the lemma proposed fixes divergence but is too difficult to be proved automatically, even with the assistance of the divergence critic. See example 34 at the end of this section for more details. Example 25 is discussed in more detail in the related work in Section 9 as it demonstrates the superiority of difference matching over generalization techniques for divergence analysis. Examples 26 to 28 require little discussion. Finally, examples 29 and 30 demonstrate that the critic can cope with divergence in moderately complex theories containing conditional equations.

The results are very pleasing. Using the divergence critic, the 30 theorems listed (with the exception of 24) can all be proved from the definitions alone. To provide an indication of the difficulty of these theorems, the Nqthm system (Boyer & Moore, 1979), which is perhaps the best known explicit induction theorem prover, was unable to prove more than half these theorems from the definitions alone. To be precise, Nqthm failed on 5, 6, 7, 8, 9, 11, 12, 13, 14, 15, 18, 19, 21, 22, 24, 25, 26, 27 and 28. Of course, with the addition of some simple lemmas, Nqthm is able to prove all these theorems. Indeed, in many cases, Nqthm needs the same lemmas as those proposed by the divergence critic and required by Spike. This suggests that the divergence critic is not especially tied to the particular prover used nor even to the implicit induction setting.

To test this hypothesis, I presented the output of a diverging proof attempt from Nqthm to the critic. I chose the commutativity of multiplication as this is perhaps the simplest theorem which causes Nqthm to diverge. The critic proposed the lemma,

```
(EQUAL (TIMES Y (ADD1 X)) (PLUS Y (TIMES Y X)))
```

where `TIMES` and `PLUS` are primitives of Nqthm's logic recursively defined on their first arguments. This is exactly the lemma needed by Nqthm to prove the commutativity of multiplication. Nqthm fails on many of the other examples for similar reasons to Spike, and divergence analysis identifies an appropriate lemma. This supports the suggestion that the divergence critic is likely to be useful for a wide variety of provers.

The divergence critic has several limitations. Recognizing divergence is, in general, undecidable since it reduces to the halting problem. The divergence critic will therefore sometimes fail to identify a diverging proof attempt. In addition, the critic will sometimes identify a "divergence" pattern when the proof attempt is not diverging. Even when divergence is correctly identified, the critic will sometimes fail to speculate an appropriate lemma. Finally, the critic only speculates wave-rules. Whilst many theories contain a large number of wave-rules, and these are often very useful for fixing divergence, other types of lemma can be needed.

Table 2 lists four theorems on which the divergence critic fails. These problems are representative of the different ways in which the critic can fail. The two main cause of failure are overlapping divergence patterns, and the inability of the heuristics to speculate an appropriate right hand side for a lemma. Again times are those to speculate lemmas and not to find a proof of the theorem.

Example 31 is a commuted version of the recursive definition of multiplication ($\times$ is defined recursively on its second argument position). Spike's attempt to prove this theorem

| No | Theorem | Lemmas speculated | Time/s |
|---|---|---|---|
| 31 | s(x) × y=y+(x × y) | – | n/a |
| 32 | x × y=y × x | s(X)+Y=X+s(Y) | 8.0 |
| 33 | x+(y+(z+(v+w))) = w+(x+(y+(z+v))) | s(X)+Y=s(X+Y)<br>s(X)+Y=X+s(Y) | 17.7 |
| 34 | qrev(qrev(x,[y]),z)=y :: qrev(qrev(x,[]),z) | qrev(Y,X :: Z)) = qrev(X :: Y,Z)<br>qrev(qrev(X,Y :: Z),W)=qrev(qrev(Y :: X,Z),W)<br>qrev(qrev(X,Y :: [Z]),W)=Z :: qrev(qrev(X,[Y]),W) | 9.6 |

Table 2: Some of the divergence critic's failures.

diverges, generating the following sequence of equations,

$$\begin{aligned} s(y) + (x + (x \times y)) &= s(y) + (y + (x \times y)) \\ s(s(y)) + (x + (x + (x \times y))) &= s(x) + (s(y) + (x + (x \times y)) \\ s(s(s(y))) + (x + (x + (x + (x \times y)))) &= s(x) + (s(s(y)) + (x + (x + (x \times y)))) \\ &\vdots \end{aligned}$$

Divergence analysis of the left hand sides of these equations suggests the need for a rule of the form,

$$\boxed{s(\underline{Y})} + \boxed{(X + \underline{Z})} = \boxed{F(\underline{Y + Z})}$$

Unfortunately the heuristics for lemma speculation are not sufficiently strong to suggest a suitable instantiation for $F$ (for example, $\lambda z \,.\, s(X + z)$). This lemma is rather complex and is the result of two overlapping divergence patterns. If the annotations are considered separately, they suggest the rules,

$$\begin{aligned} \boxed{s(\underline{X})} + Y &= \boxed{s(\underline{X + Y})} \\ Y + \boxed{(X + \underline{Z})} &= \boxed{X + (\underline{Y + Z})} \end{aligned}$$

With these two rules, SPIKE finds a proof without difficulty.

Example 32 is the commutativity of multiplication. The divergence critic identifies a divergence pattern and proposes the transverse wave rule,

$$\boxed{s(\underline{X})} + Y = X + \boxed{s(\underline{Y})}$$

However, SPIKE is unable to prove the commutativity of multiplication with the addition of this rule. The proof attempt is now somewhat simpler and contains the diverging sequence of equations,

$$\begin{aligned} x + (y + (x + (x * y))) &= y + (x + (x + (x * y))) \\ x + (y + \boxed{(x + (\underline{x + (x * y)}))}) &= y + \boxed{(x + (\underline{x + (x + (x * y))}))} \\ x + (y + \boxed{(x + (\underline{x + (x + (x * y))}))}) &= y + \boxed{(x + (\underline{x + (x + (x + (x * y)))}))} \\ &\vdots \end{aligned}$$

Unfortunately the heuristics for instantiating the right hand side of speculated lemmas are not strong enough to suggest the rule,

$$X + \boxed{(Y + \underline{Z})} = \boxed{Y + \underline{(X + Z)}}$$

With this rule, Spike finds a proof of the commutativity of multiplication without difficulty. The difficulties in speculating this rule arise because the wave-front is stuck in a similar position on both sides of the equality. There are few clues therefore to suggest how to ripple it up to the top of the term tree.

In example 33, the divergence critic proposes a lemma where one is not needed. Spike is able to find a proof of this theorem from the definitions alone using 16 inductions. Three of these inductions are on the equations,

$$\begin{aligned} 0 + x &= x \\ \boxed{s(\underline{0})} + x &= \boxed{s(\underline{x})} \\ \boxed{s(\underline{s(0)})} + x &= \boxed{s(\underline{s(x)})} \end{aligned}$$

This sequence of equations satisfies the divergence critic's preconditions. The critic therefore proposes wave rules for moving accumulating successor functions off the first argument position of $+$. Although the proposed lemmas are not necessary, either give a much shorter and simpler proof needing just 7 inductions.

Example 34 is the lemma speculated in example 24. Divergence analysis of Spike's attempt to prove this theorem identifies term structure accumulating on the second (alias accumulator) argument of *qrev*. The first two lemmas proposed for removing this term structure are of no use as they are subsumed by the recursive definition of *qrev*. The third lemma also fails to prevent divergence. This lemma simplifies two element lists in the second argument position of *qrev*. However, divergence will still occur as the prover cannot simplify lists that occur in the second argument position of *qrev* which contain 3 or more elements. Divergence can be overcome if we introduce a derived function for appending onto the end of a list. This can be used to simplify terms in which a list of arbitrary size occurs on the second argument position of *qrev*. For example, we can simplify with the rule,

$$qrev(X, Y) = app(qrev(X, nil), Y)$$

Unfortunately, append does not occur in the specification of the theorem so it is difficult to find a heuristic that would speculate such a rule.

## 9. Related Work

Critics for monitoring the construction of proofs were first proposed by Ireland for the Clam prover (Ireland, 1992). In this framework, failure of one of the proof methods automatically invokes a critic. Various critics for explicit induction have been developed that speculate missing lemmas, perform generalizations, look for suitable case splits, *etc.* As rippling plays a central role in Clam's proof methods, many of the heuristics are similar to those described here (Ireland & Bundy, 1992). There are, however, several significant differences. First, the divergence critic described here works in an implicit (and not an explicit) induction setting. Second, the divergence critic is not automatically invoked but must identify when the proof is failing. Third, the divergence critic is less specialized. These last two differences reflect the fact that critics in Clam are usually associated with the failure of a particular precondition to a heuristic. The same divergence pattern can, by comparison, arise for many different reasons: the need to generalize variables apart, to generalize common subterms, to add a lemma, *etc.* Fourth, the divergence critic must use difference matching to annotate terms; in Clam, terms are usually already appropriately annotated. Finally, the divergence critic is less tightly coupled to the the theorem prover's inference rules or heuristics. The critic can therefore exploit the strengths of the prover without needing to reason about the complex rules or heuristics being used. For instance, the divergence critic has no difficulty identifying divergence in complex situations like nested or mutual inductions. The critic also benefits from the powerful simplification rules used by Spike.

Divergence has been studied quite extensively in completion procedures. Two of the main novelties of the critic described here are the use of difference matching to identify divergence, and the use of rippling in the speculation of lemmas to overcome divergence. Dershowitz and Pinchover, by comparison, use most specific generalization to identify divergence patterns in the critical pairs produced by completion (Dershowitz & Pinchover, 1990). Kirchner uses generalization modulo an equivalence relation to recognise such divergence patterns (Kirchner, 1987); meta-rules are then synthesized to describe infinite families of rules with some common structure. Thomas and Jantke use generalization and inductive inference to recognize divergence patterns and to replace infinite sequences of critical pairs by a finite number of generalizations (Thomas & Jantke, 1989). Thomas and Watson use generalization to replace an infinite set of rules by a finite complete set with an enriched signature (Thomas & Watson, 1993).

Generalization modulo an equivalence enables complex divergence patters to be identified. However, it is in general undecidable. Most specific generalization, by comparison, is more limited. It cannot recognize divergence patterns which give nested wave-fronts like,

$$\boxed{s(\underline{\underline{\boxed{s(\underline{x})} + x}})}.$$

In addition, most specific generalization cannot identify term structure in wave-holes. For example, consider the divergence sequence of equations produced when Spike attempts to prove example 25 from Section 8,

$$\begin{aligned} rev(qrev(x, nil)) &= x \\ rev(qrev(x, cons(y, nil))) &= cons(y, x) \\ rev(qrev(x, cons(z, cons(y, nil)))) &= cons(z, cons(y, x)) \\ &\vdots \end{aligned}$$

Divergence analysis identifies term structure accumulating within the accumulator argument of $qrev$,

$$\begin{array}{rcl} rev(qrev(x, nil)) & = & x \\ rev(qrev(x, \boxed{cons(y, \underline{nil})})) & = & \boxed{cons(y, \underline{x})} \\ rev(qrev(x, \boxed{cons(z, \underline{cons(y, nil)})})) & = & \boxed{cons(z, \underline{cons(y, x)})} \\ & \vdots & \end{array}$$

This annotated sequence is the unique maximal difference match. These annotations suggest the need for the wave rule,

$$rev(qrev(X, \boxed{cons(Y, \underline{nil})})) = \boxed{cons(Y, \underline{rev(qrev(X, nil))})}.$$

This rule allows the proof to go through without divergence. By comparison, most specific generalization seems to be unable to identify this rule. The most specific generalization of the left hand side of this sequence gives the term $rev(qrev(X, Z))$ (or, ignoring the first term in the sequence, $rev(qrev(X, cons(Y, Z)))$). Most specific generalization cannot, however, identify the more useful pattern, $rev(qrev(X, cons(Y, nil)))$.

Nqthm contains a simple test for divergence based on subsumption. For instance, on example 13 of the last section, Nqthm is unable to simplify the following subgoal in the step case of the proof,

```
(EQUAL (ROT (LENGTH X) (APPEND X (LIST Z)))
       (CONS Z (ROT (LENGTH X) X))))
```

Note that this is the lemma speculated by the divergence critic. Nqthm generalizes `(LENGTH X)` in this subgoal giving the false conjecture,

```
(EQUAL (ROT Y (APPEND X (LIST Z)))
       (CONS Z (ROT Y X))))
```

After several more attempts at induction and generalization, Nqthm realizes the proof is diverging since a subgoal is subsumed by its parent. As the proof is therefore about to loop, Nqthm gives up. No attempt is made to analyse the failed proof attempt to identify where it started to go wrong. In addition, subsumption is a very weak test for divergence, much weaker than tests based on difference matching or generalization. This subsumption test recognizes divergence on just a small number of the failed examples in the last section.

## 10. Conclusions

I have described a divergence critic, a computer program which attempts to identify diverging proof attempts and to propose lemmas and generalizations which overcome the divergence. The divergence critic has proved very successful; it enables the system Spike to prove many theorems from the definitions alone. The divergence critic's success can be largely attributed to the power of the rippling heuristic. This heuristic was originally developed for proofs using explicit induction but has since found several other applications. Difference matching is used to identify accumulating term structure which is causing divergence. Lemmas and generalizations are then proposed to ripple this term structure out of the way. There are other types of divergence which could perhaps be recognized by the divergence critic. Further research is needed to identify such divergence patterns, isolate their causes and propose ways of fixing them. This research may take advantage of the close links between divergence patterns and particular types of generalization. For instance, it may be possible to identify specific divergence patterns with the need to generalize common subterms in the theorem being proved.

## Acknowledgements

This research was supported by a Human Capital and Mobility Postdoctoral Fellowship. I wish to thank: Adel Bouhoula and Michael Rusinowitch for their invaluable assistance with SPIKE; Pierre Lescanne for inviting me to visit Nancy where most of this research was performed; David Basin, Alan Bundy, Miki Hermann, Andrew Ireland, and Michael Rusinowitch for their comments and questions; the reviewers for their comments and suggestions; the members of the Eureca and Protheo groups at INRIA; and the members of the DReaM group at Edinburgh and the MRG groups in Trento and Genova.